\documentclass[preprint,12pt]{elsarticle}

\usepackage{lineno}
\modulolinenumbers[5]
\usepackage{amssymb}
\usepackage{subfigure}
\usepackage{graphicx}
\usepackage{latexsym}
\usepackage{booktabs}
\usepackage{amsmath}
\usepackage[mathcal]{eucal}
\usepackage{multirow}
\usepackage{bm}
\usepackage[linesnumbered,ruled]{algorithm2e}
\usepackage{enumerate}
\usepackage{enumitem}
\setlist[enumerate,1]{label=(\alph*)}

\begin{document}

\begin{frontmatter}

\title{MOBA: A multi-objective bounded-abstention model for two-class cost-sensitive problems}
\author{Hongjiao Guan}

\address{School of Computer Science and Technology, Harbin Institute of Technology, Harbin, 150001, China}

\begin{abstract}
Abstaining classifiers have been widely used in cost-sensitive applications to avoid ambiguous classification and reduce the cost of misclassification. Previous abstaining classification models rely on cost information, such as a cost matrix or cost ratio. However, it is difficult to obtain or estimate costs in practical applications. Furthermore, these abstention models are typically restricted to a single optimization metric, which may not be the expected indicator when evaluating classification performance. To overcome such problems, a multi-objective bounded-abstention (MOBA) model is proposed to optimize essential metrics. Specifically, the MOBA model minimizes the error rate of each class under class-dependent abstention constraints. The MOBA model is then solved using the non-dominated sorting genetic algorithm II, which is a popular evolutionary multi-objective optimization algorithm. A set of Pareto-optimal solutions will be generated and the best one can be selected according to provided conditions (whether costs are known) or performance demands (e.g., obtaining a high accuracy, F-measure, and etc). Hence, the MOBA model is robust towards variations in the conditions and requirements. Compared to state-of-the-art abstention models, MOBA achieves lower expected costs when cost information is considered, and better performance-abstention trade-offs when it is not.
\end{abstract}

\begin{keyword}
Abstaining classification \sep Cost-sensitive problems \sep Multi-objective optimization (MOO) \sep Evolutionary algorithm (EA) 
\end{keyword}

\end{frontmatter}

%\linenumbers

\section{Introduction}
Forcing the classification of uncertain instances in safety-critical applications can lead to misclassification, which can result in economic losses or increased costs. In contrast, the cost of possible errors can be reduced by abstaining from ambiguous classification, which has been used in cost-sensitive fields \cite{chen2018evidential,kang2017reliable,lin2018biomedical,wang2017fault}.  

The general classification rule with reject option in binary classification is shown in Eq.~(\ref{eq1}) and can be explained as follows. If the confidence score $s$ belonging to the positive class is larger than $t_2$, then example $m$ is classified as positive (+); if $s$ is not larger than $t_1$, then $m$ is classified as negative (-); otherwise, the example is not labeled (rejected, R). 
\begin{equation}\label{eq1}
C_{t_1,t_2}(m) = \begin{cases}
+,  & \text{if } s(m)>t_2;  \\
-, & \text{if } s(m) \leq t_1;   \\
R, & \text{otherwise}.
\end{cases}
\end{equation}
When $t_1=t_2$, Eq.~(\ref{eq1}) reverts to the traditional binary classification rule. The rejection thresholds $t_1$ and $t_2$ define the decision boundaries and the task of training the abstention classifier lies in determining the two rejection thresholds, which can be enforced by establishing different abstention models. Note that only binary classification is discussed in this paper.

In the context of abstaining classification, the ideal situation is to establish a cost minimization model, which requires the costs of correct classification, misclassification, and rejection to be known. However, it is often difficult to obtain or estimate cost information in many real-world classification problems. For instance, in the diagnosis of normal and cancerous tissues, the imaging characteristics of some tissues are ambiguous, in which case it is hard to make a definitive diagnosis. Both misdiagnosis and a missed diagnosis can lead to physical and mental pain, and their costs are usually unknown and unequal. There are mainly two abstention models as follows: unconditional optimization model of the expected cost and conditional optimization model with rejection or performance constraints.

Chow \cite{chow1970optimum,chow1957optimum} expanded classical Bayesian decision theory by proposing a generalized decision theory with reject option. The well-known Bayesian decision rules are based on a minimum error rate or a minimum risk function. For example, minimum risk Bayesian classification only considers class-dependent misclassification costs. In contrast, once the reject option is added, the generalized Bayesian decision theory minimizes the expected risk based on the rejection costs as well as the costs of correct and incorrect classification. Tortorell \cite{tortorella2004reducing,tortorella2000optimal} proposed the following implementable abstention model based on the generalized theory: 
\begin{equation} \label{eq2}
\min_{t_1,t_2} \hspace{0.5em} cost(t_1,t_2),
\end{equation}
where 
\begin{align}  \label{eq3}
&cost(t_1,t_2)=  \notag \\
&p(+)\cdot CFN \cdot fnr(t_1)+p(-) \cdot CTN \cdot tnr(t_1)+ \notag \\
& p(+)\cdot CTP \cdot tpr(t_2) + p(-) \cdot CFP \cdot fpr(t_2)+ \\
& p(+)\cdot CRP \cdot rpr(t_1,t_2) + p(-) \cdot CRN \cdot rnr(t_1,t_2), \notag
\end{align}
where $p(+)$ and $p(-)$ are the prior probabilities of the positive and negative classes, respectively, CFN represents the cost of a false negative, and $fnr$ denotes the ratio of false negative examples among all positive examples. Likewise for CTN, CTP, CFP, $tnr$, $tpr$, and $fpr$. CRP and CRN indicate the rejection costs for positive and negative classes, respectively, and $rpr$ and $rnr$ are the ratios of rejected examples with respect to the positive and negative classes, respectively. This abstention model requires the complete cost information including the costs of correct classification, misclassification, and rejection to be known. 

Pietraszek \cite{pietraszek2007use} proposed a bounded-abstention (BA) model that adds an abstention constraint and only requires knowledge of the misclassification costs. The BA model can be represented as:
\begin{equation} \label{eq4}
\begin{split}  
&\min_{t_1,t_2} \hspace{0.5em}
\frac{CFN \cdot FN(t_1)+CFP \cdot FP(t_2)}{TN(t_1)+FP(t_2)+TP(t_2)+FN(t_1)},  \\
&s.t.\hspace{0.5em} rej(t_1,t_2) \leq k_{max},
\end{split}
\end{equation}
where $FN$ ($FP$) refers to the number of false negatives (positives); $TN$ ($TP$) represents the number of true negatives (positives); and $rej$ denotes the overall reject rate, i.e., the number of rejected examples divided by the sample size. It is useful to define a cost ratio of CFN to CFP. When the values of CFN and CFP are the same, i.e., the cost ratio is 1, model~(\ref{eq4}) is to minimize the error rate under an abstention constraint. In~\cite{vanderlooy2009roc}, the receiver operating characteristic (ROC) isometric model is employed to minimize the reject rate under class-dependent performance constraints. In the performance constraints, the class distribution and misclassification cost ratios are considered. Although the BA and ROC isometric models avoid setting complete costs (CTP/N, CFP/N, and CRP/N), misclassification costs or their cost ratio are still required.

The abstention models mentioned above have the following shortcomings:
\begin{itemize}
\item they require cost information to be known. However, in practical problems: (a) such costs are hard to obtain or estimate; (b) they are usually dependent on classes and are asymmetric; and (c) in some cases, the cost information evolves over time, which prevents use of the trained model during the test phase. 
\item they only optimize a single performance metric, such as the expected cost, error rate, precision, or F-measure. The optimized metric may not be useful when practitioners evaluate the classification performance. 
\end{itemize}

To overcome these drawbacks, a multi-objective bounded-abstention (MOBA) model is proposed that minimizes two essential metrics, namely, the false positive rate and false negative rate, under class-dependent abstention constraints. The MOBA model optimizes essential metrics, via which any simple or complicated metric can be calculated as long as the sample sizes of two classes are fixed. In addition, the MOBA model is applicable regardless of whether the costs are known or unknown. A popular evolutionary algorithm called the non-dominated sorting genetic algorithm II (NSGA-II) \cite{deb2002fast} is used to solve the multi-objective optimization (MOO) problem with constraints. 

The remainder of the paper is organized as follows. Section~\ref{sec2} introduces the motivation and optimization method of the proposed MOBA model, discusses the methods of selecting the best abstaining classifier under different conditions, and summarizes the advantages of the MOBA model. The results of two experiments, one with costs and one without, are presented in Section~\ref{sec4}, and the conclusions drawn from this research are summarized in Section~\ref{sec5}.

\section{Proposed MOBA model} \label{sec2}
% Table 1
\begin{table}[!htb]
  \centering
  \caption{Confusion matrix with rejection in binary classification problems}
           \begin{tabular}{|l|c|c|c|c|}
    \toprule
    \multicolumn{2}{|c|}{\multirow{2}{*}{\makebox[7em]{Confusion matrix}}} & \multicolumn{3}{c|}{\makebox[10.5em]{Predicted label}} \\
\cmidrule{3-5}     \multicolumn{2}{|c|}{} & \makebox[3.5em]{+}   & \makebox[3.5em]{-}   & \makebox[3.5em]{R} \\
    \midrule
    \multirow{2}{*}{\makebox[3.5em]{Real label}} & \makebox[3.5em]{+}   & \makebox[3.5em]{TP}  & FN  & RP \\
\cmidrule{2-5}        & \makebox[3.5em]{-}   & FP  & TN  & RN \\
    \bottomrule
    \end{tabular}%
  \label{t1}%
\end{table}%
In this section, the motivation that inspires to propose the MOBA model is first presented in Section~\ref{sec2.1} and then several basic concepts related to MOO problems are provided in Section~\ref{sec2.2}. Next, the NSGA-II algorithm is introduced along with the implementation details of solving the MOBA model (Section~\ref{sec2.3}). Selection of the best abstaining classifier and the advantages of the MOBA model are explained in Sections~\ref{sec2.4} and~\ref{sec2.5}. 
 
\subsection{Motivation} \label{sec2.1}

Regardless of whether the expected cost, error rate, or F-measure are to be optimized, previous abstention models have always required certain essential metrics to be provided, such as the true positive/negative rate ($tpr/tnr$), false positive/negative rate ($fpr/fnr$), positive/negative predictive value ($ppv/npv$), and rejected positive/negative rate ($rpr/rnr$). For example, when minimizing the average cost in (\ref{eq3}), six essential metrics ($tpr/tnr$, $fpr/fnr$, and $rpr/rnr$) are required while when maximizing the F-measure, $ppv$ and $tpr$ are needed to be known. All the essential metrics can be calculated from the confusion matrix with rejection (Table~\ref{t1}). For example, $rpr=RP\,/\,(TP+FN+RP)$, i.e., the number of rejected positive examples (RP) divided by the number of all positive examples (TP+FN+RP).

In view of this, optimizing the essential metrics is a natural and reasonable idea. When the sample sizes of two classes are fixed, the confusion matrix with rejection has four degrees of freedom. Therefore, to obtain a definite rejection classifier, four essential metrics should be determined. The MOBA model optimizes the false positive and negative rates under the constraints of the rejected positive and negative rates. A formal description of the MOBA model is as follows:
\begin{align} 
%\begin{split}  
&\min_{t_1,t_2} \hspace{0.5em}
\bm{F(t)}=(F_1(\bm{t}),F_2(\bm{t}))=(fpr(t_2),fnr(t_1)), 
\label{eq5}\\
& s.t.\hspace{0.5em} 
\begin{cases} \label{eq6}
rpr(\bm{t}) \leq p_{max}, \\
rnr(\bm{t}) \leq n_{max}, \\
t_1<t_2,
\end{cases}
%\end{split}
\end{align}
where $\bm{t}=(t_1,t_2)$ is the decision vector, $\bm{F(t)}$ represents the objective function vector, which contains two objective functions $fpr$ (related to $t_2$) and $fnr$ (related to $t_1$), and $t_1$ and $t_2$ ($t_1 < t_2$) denote the rejection thresholds. Note that the definitions of $fpr$ and $fnr$ here are different from those in Eq.~(\ref{eq3}). In this case, $fpr$ ($fnr$) denotes the ratios of false positive (negative) examples among the classified negative (positive) examples. After the two rejection thresholds have been determined, $fpr=FP\,/\,(TN+FP)$ and $fnr=FN\,/\,(TP+FN)$ can be calculated. The feasible solution set $T\subseteq \mathbb{R}^2$ is the set of decision vectors that satisfy the constraints: 
\begin{equation} 
T=\{\bm{t}\in \mathbb{R}^2| rpr(\bm{t}) \leq p_{max}, rnr(\bm{t}) \leq n_{max}, t_1 < t_2 \},
\end{equation}
where $p_{max}$ and $n_{max}$ are the hyperparameters. Note that maximizing $tpr$ and $tnr$ is equivalent to minimizing $fpr$ and $fnr$ in Eq.~(\ref{eq5}).

\subsection{Concepts associated with MOO problems} \label{sec2.2}
Since MOO problems involve multiple conflicting objectives, the comparison relation in single-objective optimization problems is not applicable. For a given decision vector, some objectives are optimal whereas others are not, and optimizing the suboptimal objectives may degrade the optimal objectives. The partially ordering relation, i.e., the Pareto dominance, is used to compare decision vectors in MOO problems. A decision vector $\bm{a} \in T$ is said to Pareto dominate another decision vector $\bm{b} \in T$, denoted as $\bm{a}\prec \bm{b}$, if and only if (\textit{iff}): 
\begin{equation} \label{eq7}
\forall i \in \{1,2\}, F_i(\bm{a}) \leq F_i(\bm{b}) \wedge \exists j \in \{1,2\}, F_j(\bm{a}) < F_j(\bm{b}).
\end{equation}
A decision vector $\bm{t}$ is non-dominated with regard to $T$ \textit{iff} there is no decision vector in $T$ that dominates $\bm{t}$. Such non-dominated solutions are referred to as Pareto-optimal. The set of Pareto-optimal solutions related to $T$ is referred to as the Pareto-optimal set (POS):
\begin{equation} 
POS = \{\bm{t} \in T | \neg \exists  \bm{t'} \in T, \bm{t'} \prec \bm{t} \}.
\end{equation}
The set of objective vectors corresponding to the POS is referred to as the Pareto-optimal front (POF):
\begin{equation} 
POF  = \{\bm{F(t)} \in [0,1]^2 | \bm{t} \in POS\}.
\end{equation}

Evolutionary algorithms (EAs) based on Pareto dominance exhibit excellent performance when solving MOO problems with few (two or three) objectives \cite{zitzler2000comparison}. EAs search the set of Pareto-optimal solutions in parallel in a single run. Popular EAs include the NSGA-II \cite{deb2002fast}, strength Pareto evolutionary
algorithm 2 (SPEA2) \cite{zitzler2001spea2}, and Pareto envelope based
selection algorithm II (PESA-II) \cite{corne2001pesa}. The popular NSGA-II algorithm was adopted in this study to solve the proposed MOBA model. 

\subsection{Evolutionary MOO of MOBA} \label{sec2.3}
In this section, the NSGA-II algorithm is introduced along with the details required to optimize the proposed MOBA model. NSGA-II improves on the previous NSGA \cite{srinivas1994muiltiobjective} by developing a fast non-dominated sorting and elitism strategy. The basic framework of NSGA-II is presented in Algorithm 1.
\begin{algorithm}[!htb]
\caption{The basic framework of NSGA-II.} 
   \SetAlgoNoLine
    \KwIn{$popsize$, the population size; $gensize$, the max generation number\\}
    \KwOut{$P$, the population at the end of $gensize$ generations \\}
    \BlankLine
$P^0 \leftarrow$ pop-initialization($popsize$);\\
$t \leftarrow$ 0;\\
\While{$t< gensize$}{
$Q^t \leftarrow $ make-new-pop($P^t$);\\
$R^t \leftarrow P^t \cup Q^t$;\\
$P^{t+1} \leftarrow$ elite-preservation($R^t, popsize$);\\
$t \leftarrow t+1$;}
Return $P^{t+1}$.
\BlankLine
\textbf{Function} make-new-pop($P^t$)\\
${\mathcal{F}}^t \leftarrow$ non-dominated-sort($P^t$);\\
${\mathcal{S}}^t \leftarrow$ crowding-distance-assignment(${\mathcal{F}}^t$);\\
$parent \leftarrow$ tournament-selection($P^t, {\mathcal{F}}^t, {\mathcal{S}}^t, popsize$);\\
$Q^t \leftarrow$ genetic-operation($parent$);\\
Return $Q^t$.
\BlankLine
\textbf{Function} elite-preservation($R^t, popsize$)\\
 ${\mathcal{F}}^t \leftarrow$ non-dominated-sort($R^t$);\\
$P^{t+1} \leftarrow \varnothing$; $i \leftarrow$ 0;\\
\While{$|P^{t+1}|+|{\mathcal{F}_i^t}| \leq popsize$}
{
$P^{t+1} \leftarrow P^{t+1} \cup {\mathcal{F}_i^t}$;\\
$i \leftarrow i+1$; 
}
${\mathcal{S}_i^t} \leftarrow$ crowding-distance-assignment(${\mathcal{F}_i^t}$);\\
$P^{t+1} \leftarrow P^{t+1} \cup {\mathcal{F}_i^t}[\{{\mathcal{S}_i^t}\}_1^{popsize-|P^{t+1}|}]$;\\
Return $P^{t+1}$.
\end{algorithm}

\textbf{Pop-initialization} generates the initial population that includes $popsize$ individuals (chromosomes). Each decision vector $\bm{t}$ denotes an individual or a chromosome. In the MOBA problem, the two variables in each decision vector, i.e., the two rejection thresholds $t_1$ and $t_2$, are encoded with real values. Specifically, the scores $s$ of the training examples are first determined via a scoring classifier. Traditional classification methods, such as support vector machine and k-nearest neighbor, can be used as the scoring classifier \cite{fawcett2006introduction}. Let $s_{min}$ and $s_{max}$ denote the minimal and maximal ones among the scores of all training examples, respectively. Then, $t_1$ and $t_2$ are randomly generated in the range of [$s_{min}, s_{max}$] only if $t_1 < t_2$ is satisfied. In this study, $popsize$ and $gensize$ are set to 20 and 100, respectively.

\begin{figure}[!htb]
\centering
\includegraphics[width=0.68\textwidth]{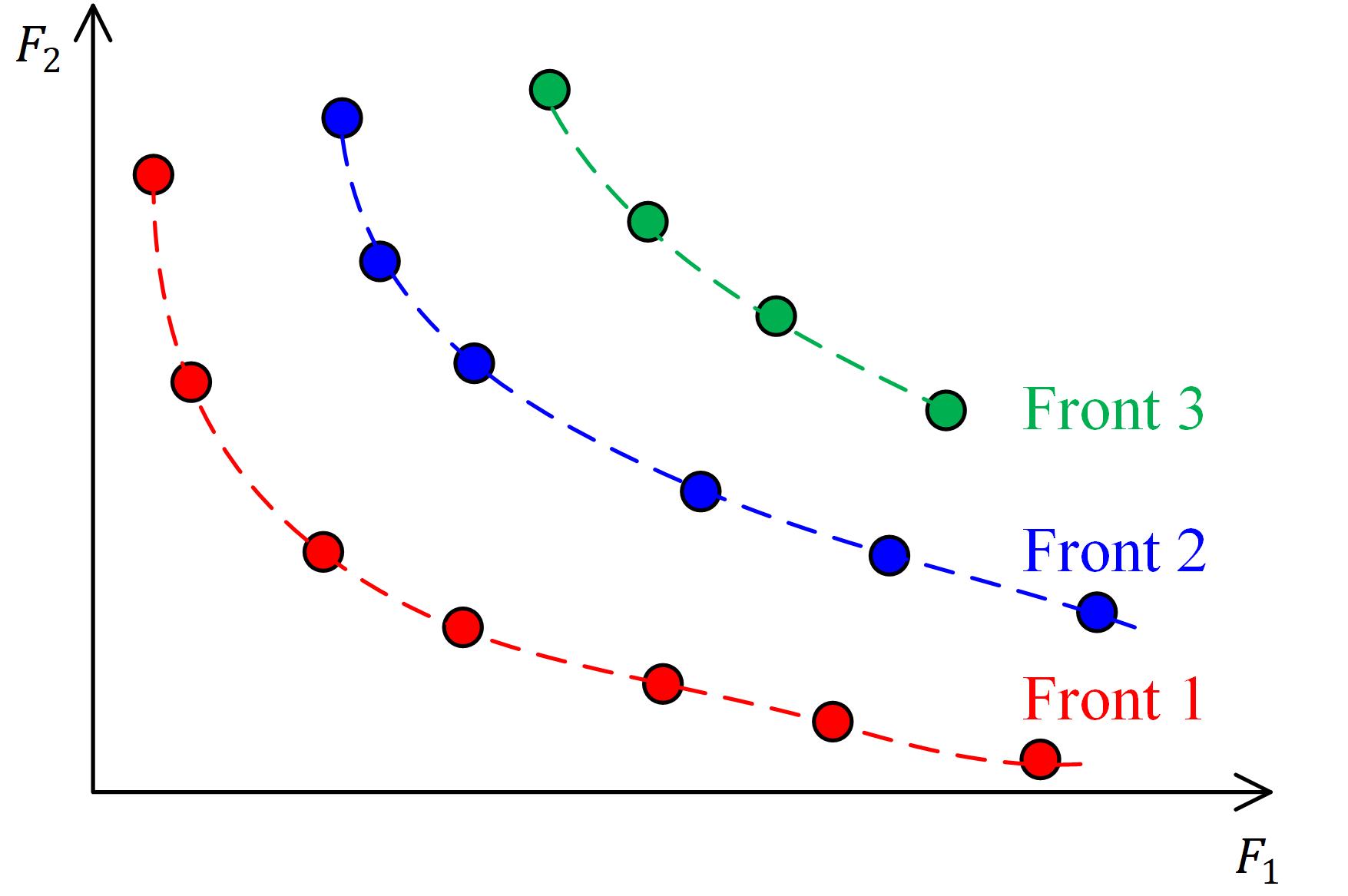}
\caption{The illustration of front sets}\label{f1}
\end{figure}

\textbf{Non-dominated-sort} sorts the individuals in population $P^t$ based on non-domination and outputs the front set ${\mathcal{F}}^t=\{{\mathcal{F}}_1^t, {\mathcal{F}}_2^t, \cdots, {\mathcal{F}}_n^t\}$. The individuals in each front ${\mathcal{F}}_i^t$ ($i \in \{1, 2, \cdots, n\}$) are non-dominated while the individuals belonging to ${\mathcal{F}}_i^t$ are dominated by the individuals in front ${\mathcal{F}}_j^t$ ($j<i, j \in \{1, 2, \cdots, n\}$) in the $t^{th}$ generation. The front sets are depicted in Figure~\ref{f1}, where $F_1$ and $F_2$ represent the values of two objective functions. For the detailed sort algorithm, please refer to~\cite{deb2002fast}. The objective values $fpr$ and $fnr$ are computed using the validation set. Specifically, all examples are divided into three parts consisting of the training, validation, and test sets. The training set is employed to construct a scoring classifier, and the scores of the examples in the validation set are computed using the scoring classifier. Given $t_1$ and $t_2$ (corresponding to the variables in each individual), Eq.~(\ref{eq1}) can be applied to the validation examples, thereby allowing the basic metrics in Table~\ref{t1} to be calculated. If the constraints~(\ref{eq6}) are not satisfied, $fpr$ and $fnr$ are assigned the maximum value of one, in which case the corresponding individual is not considered.

\begin{figure}[!htb]
\centering
\includegraphics[width=0.68\textwidth]{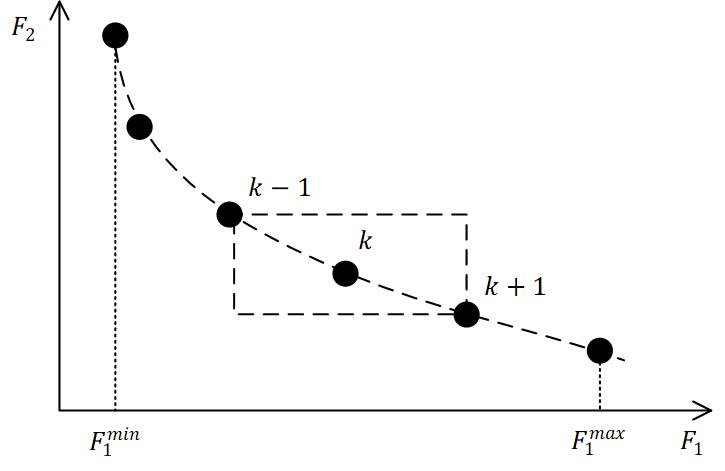}
\caption{The illustration of the crowding distance}\label{f2}
\end{figure}
\textbf{Crowding-distance-assignment} computes the crowding distances of individuals in each front and outputs the spread set ${\mathcal{S}}^t=\{{\mathcal{S}}_1^t, {\mathcal{S}}_2^t, \cdots, {\mathcal{S}}_n^t\}$, where ${\mathcal{S}}_i^t$ ($i \in \{1, 2, \cdots, n\}$) stores the crowding distances of the individuals in front ${\mathcal{F}}_i^t$ in the $t^{th}$ generation. Note that since the crowding distance is assigned within each front, it is meaningless to compare the crowding distances of two individuals from different fronts. The essential idea of calculating the crowding distance is to sort all individuals in the same front based on each objective function and to average the Euclidean distances between the nearest neighbors in each dimension of the objective function space. An illustration of the crowding distance is shown in Figure~\ref{f2}. For objective function $F_1$, the distance $d_1$ of the $k^{th}$ individual is:
\begin{equation} 
d_1 = \frac{F_1^{k+1}-F_1^{k-1}}{F_1^{max}-F_1^{min}},
\end{equation}
where $F_1^{k+1}$ and $F_1^{k-1}$ are the objective values of the $(k+1)^{th}$ and $(k-1)^{th}$ individuals in dimension $F_1$, respectively, and $F_1^{max}$ and $F_1^{min}$ are the maximum and minimum values in the front, respectively. A similar approach can be used to obtain the distance of the $k^{th}$ individual in other dimensions of the objective function space. Finally, the crowding distance of the $k^{th}$ individual is the sum of the distances in all dimensions. Note that the crowding distance of the boundary individual is assigned as infinite. In each ${\mathcal{S}}_i^t$, the crowding distances are sorted in descending order.

\textbf{Tournament-selection} uses a binary tournament strategy to select $popsize$ individuals from the population $P^t$ as follows. First, two individuals are randomly selected from $P^t$. Then, one of them is chosen based on two criteria, namely, the front rank and crowding distance. If the two individuals are in different fronts, the individual with the lower front rank is selected. However, if the front ranks of the two individuals are the same, then the individual with the larger crowding distance is selected to maintain solution diversity. The binary tournament selection process is performed $popsize$ times to obtain $popsize$ individuals, which then constitute the $parent$ population.

\textbf{Genetic-operation} performs the simulated binary crossover (SBX)~\cite{agrawal1995simulated} and polynomial mutation~\cite{kakde2004survey} operations in the real value coded evolutionary MOO algorithm. Specifically, in SBX, two children chromosomes are generated as follows~\cite{beyer2001self}:
\begin{align}
y_{1,m} = \frac{1}{2}[(1-\beta_m)x_{1,m}+(1+\beta_m)x_{2,m}], \label{eq11} \\
y_{2,m} = \frac{1}{2}[(1+\beta_m)x_{1,m}+(1-\beta_m)x_{2,m}], \label{eq12}
\end{align}
where $y_{1,m}$ and $y_{2,m}$ ($m \in \{1,2\}$) are the $m^{th}$ variables in the two children $y_1$ and $y_2$, respectively; $x_{1,m}$ and $x_{2,m}$ are the $m^{th}$ variables of the randomly selected parents, respectively, and $\beta_m (\geq 0)$ is a sample from a random number generator having the density:
\begin{equation}
p(\beta) = \begin{cases}
\frac{1}{2}(\eta_c+1)\beta^{\eta_c}, &\text{if } 0 \leq \beta \leq 1\\
\frac{1}{2}(\eta_c+1)\frac{1}{\beta^{\eta_c+2}}, &\text{if } \beta>1 
\end{cases} 
\end{equation}
where $\eta_c$ is the distribution index of the crossover. The distribution can be obtained from a uniformly sampled random number $u$ between (0,1):
\begin{equation} \label{eq14}
\beta(u) = \begin{cases}
(2u)^{\frac{1}{\eta_c+1}}, & \text{if } u \leq 0.5 \\
\frac{1}{(2-2u)^{\frac{1}{\eta_c+1}}}. &\text{if } u>0.5
\end{cases} 
\end{equation}
When generating children individuals using the SBX operation, a random number $u$ between (0,1) is obtained, $\beta$ is computed as per Eq.~(\ref{eq14}), and the child variables are obtained via Eqs.~(\ref{eq11}) and (\ref{eq12}). After two child chromosomes are obtained, the constraint $t_1<t_2$ is checked. If the constraint is not satisfied, a new random number between (0,1) is generated and the computation process is repeated.

In polynomial mutation, a child chromosome is generated as follows: 
\begin{equation}  \label{eq15}
y_m = x_m+(x_m^u-x_m^l)\delta_m,
\end{equation}
where $y_m$ ($m \in \{1,2\}$) is the $m^{th}$ variable in the child chromosome $y$, $x_m$ is the $m^{th}$ variable of the parent chromosome $x$, and $x_m^u$ and $x_m^l$ are the upper and lower bounds, respectively. In the MOBA model, $x_m^u = 1$ and $x_m^l =0$, and $\delta_m$ follows the polynomial probability distribution \cite{kakde2004survey}:
\begin{equation}
p(\delta)=\frac{1}{2}(\eta_m+1)(1-|\delta|^{\eta_m}),
\end{equation}
where $\eta_m$ is the distribution index for mutation and $\delta$ can be calculated as: 
\begin{equation} \label{eq17}
\delta(u) = \begin{cases}
(2u)^{\frac{1}{\eta_m+1}}-1, & \text{if } u<0.5 \\
1-(2-2u)^{\frac{1}{\eta_m+1}}, &\text{if } u \geq 0.5
\end{cases} 
\end{equation}
where $u$ is a random number between (0,1). Similarly, to obtain a child chromosome, $u$ is generated randomly between (0,1) and $\delta$ is computed using Eq.~(\ref{eq17}). Finally, a child is generated as per Eq.~(\ref{eq15}) and the constraint $t_1<t_2$ is checked. New values of $u$ are generated until the constraint is satisfied. Finally, the offspring population $Q^t$ is produced from the chromosomes in the $parent$ population.

In the experiments, the commonly used hyperparameters in the NSGA-II algorithm are set as follows \cite{deb2002fast,agrawal1995simulated}. The crossover probability is 0.9; the mutation probability is $1/v$, where $v$ is the number of decision variables, and here, $v=2$; and the distribution indexes of both the crossover and mutation operators are set to 20. 

\textbf{Elite-preservation} selects the first $popsize$ best individuals from the combined population $R^t$. All individuals in $R^t$ are sorted based on non-domination. The individuals in the low fronts are better solutions than those in the high fronts. If the number of chromosomes in ${\mathcal{F}_1}$ is smaller than $popsize$, the next non-dominated set ${\mathcal{F}_2}$ is considered until the size of $P^{t+1}$ is larger than $popsize$. It is assumed that when the $i^{th}$ front set is added into $P^{t+1}$, the number of individuals exceeds $popsize$. The crowding distances of the solutions in ${\mathcal{F}_i}$ are assigned and sorted, and the individuals that have large crowding distances are added into $P^{t+1}$ until the size of $P^{t+1}$ is exactly equal to $popsize$.

The MOBA model requires two hyperparameters to be set, namely, the maximum abstention rates $p_{max}$ and $n_{max}$ with respect to the two classes. A larger abstention rate results in better performance, but increases the cost of dealing with rejected instances. Note that this trade-off always arises in abstaining classification applications. The hyper-parameters can be set depending on the requirements of the application (associated with the performance) and the resource limitations (associated with the reject rate). The hyper-parameter setting is presented in Sections~\ref{sec4.1} and \ref{sec4.2}. It is important to note that the abstention rates of the two classes can be set using different values. This ensures the MOBA model can be applied in situations where the class distribution is imbalanced and the error costs are asymmetric. Furthermore, another advantage of the MOBA model is that a set of classifiers is generated, instead of just one classifier. This provides the ability to select and change classifiers without incurring the expense of retraining. 

\subsection{Methods of selecting the best abstaining classifier} \label{sec2.4}
In the MOBA model, the NSGA-II algorithm outputs multiple Pareto-optimal vectors $(t_1, t_2)$, each of which corresponds to an abstaining classifier. For a fixed abstaining classifier, test examples can be classified or rejected according to Eq.~(\ref{eq1}), and essential metrics can be calculated. With this in mind, the methods of selecting the best classifier depend on the cost information and the metrics used to evaluate performance.
\begin{itemize}
\item If the complete costs (CTP/N, CFP/N, and CRP/N) are known, then the solution having the minimum expected cost (Eq.~(\ref{eq3})) is theoretically optimal. Specifically, at the end of the 100th generation, 20 Pareto-optimal vectors are obtained, each of which corresponds to a confusion matrix with rejection. Hence, $tpr/tnr$, $fpr/fnr$, and $rpr/rnr$ can be computed, and correspondingly, a set of 20 expected costs can be obtained as per Eq.~(\ref{eq3}), from which the smallest one is selected.
\item If the cost information is unknown, practitioners can compare the performance-abstention trade-offs of the Pareto-optimal solutions and select the best solution for a particular circumstance. For example, it is possible to preset tolerable maximum reject rates and then choose the classifier that has the best performance in terms of the accuracy, area under the ROC curve (AUC), etc. Once 20 confusion matrices with rejection are obtained, the corresponding rejected rates can be computed, and eligible Pareto-optimal vectors are selected according to the preset maximum value. Among the selected vectors, the best solution is the one that obtains a highest AUC, for example. Note that AUC is the average of $tpr$ and $tnr$ \cite{lopez2013insight}. 
\end{itemize}

\subsection{Superiority of MOBA over previous abstaining models} \label{sec2.5}
The advantages of the MOBA model are summarized as follows:
\begin{enumerate}
\item the MOBA model does not require set costs, and although the abstention constraints involve two hyperparameters, they are in the range of (0,1). This is distinctly different than the costs in previous models that take on unbounded real values. 

In the models of unconditional optimization of the expected cost Eq. (\ref{eq2}) and conditional optimization with constraints such as Eq. (\ref{eq4}), cost information is required to be provided, and unfortunately, it is usually unknown. Hence, empirical costs or cost ratios are commonly set in particular applications such as intrusion detection \cite{pietraszek2007classification} or cost models are used to evaluate statistical results (Section~\ref{sec4.1}). In such cases, costs can be set using any real value, which leads to constructing cost-dependent models. When the costs change, retraining models will take additional computation. In contrast, the MOBA model, which does not rely on cost information, only sets two abstention parameters in the range of (0,1).
 
\item the MOBA model is robust to varying conditions and demands since a set of Pareto-optimal vectors (corresponding to a set of abstaining classifiers) are generated. 

As explained in Section~\ref{sec2.4}, the optimal abstaining classifier can be selected according to the demands (e.g., obtaining a maximum F-measure or a minimum error rate under fixed abstentions) or conditions (whether the costs are known). Furthermore, if the costs change over time, no retraining of the MOBA model is required as a new optimal abstaining classifier can be determined simply by recalculating the expected costs.
% Table 2 data
\begin{table}[!htb]
  \centering
  \caption{Characteristics of datasets used in this study}
 \begin{tabular*}{\textwidth}{l@{\extracolsep{\fill}}cccc}
    \toprule
    Dataset & Instances & Positive & Negative & Attrsibutes \\
    \midrule
    ionosphere & 351 & 126 & 225 & 34 \\
    pima & 768 & 268 & 500 & 8 \\
    credit-g & 1000 & 300 & 700 & 20 \\
    ecoli3 & 336 & 35  & 301 & 7 \\
    hepatitis & 155 & 32  & 123 & 19 \\
    haberman & 306 & 81  & 225 & 3 \\
    cmc & 1473 & 333 & 1140 & 9 \\
    transfusion & 748 & 178 & 570 & 4 \\
    Australian & 690 & 307 & 383 & 14 \\
    \bottomrule
    \end{tabular*}%
  \label{t2}%
\end{table}%
\item compared to the BA model, the MOBA model can control the respective performance of two classes via the class-dependent abstention constraints, so the MOBA model is more applicable to imbalanced datasets or cost-sensitive problems.

The BA model has an overall reject rate constraint. When datasets are imbalanced, the reject rates of two classes may be imbalanced even though the overall reject rate constraint is satisfied. When the two abstention parameters are set using the same value, the MOBA model can avoid imbalanced reject rates because of its class-dependent abstention constraints. In addition, the two abstention parameters in the MOBA model can have different values when dealing with imbalanced datasets.
\end{enumerate}

\section{Experimental results} \label{sec4}
In the experiments, the proposed MOBA model was compared with two abstaining classification models: one considering costs (Section~\ref{sec4.1}) and the other one not (Section~\ref{sec4.2}). Table~\ref{t2} lists the datasets used in the experiments, which are available in the KEEL-dataset repository \cite{alcala2011keel}. Among these datasets, there are several ones associated with cost-sensitive classification tasks, such as \textit{pima} and \textit{credit-g}. For \textit{pima}, the task is to predict whether the patient is diabetic. The cost of missing a diabetic patient is higher than that of misdiagnosing a nondiabetic patient. \textit{Credit-g} is a dataset in the financial area, which classifies people as good or bad credit customers. It is worser to classify bad credit customers as good than the opposite case. Each dataset was divided into three distinct subsets: the training set (containing 60\% of the examples) used to generate confidence scores by training a scoring classifier, and the validation and test sets, each of which contained 20\% of the examples. The validation set was used to determine the rejection thresholds while the classification performance was evaluated via the test set. All the experimental results were obtained using MATLAB R2017a.

\subsection{Comparison of the results when the costs are considered}
\label{sec4.1}
% table 3 cost models
\begin{table*}[!htb]
  \centering
  \caption{Cost Models}
    \begin{tabular*}{\textwidth}{@{\extracolsep{\fill}}llllll}
    \toprule
        & CTP/N & CFP & CFN & CRP & CRN \\
    \midrule
    CM1 & U[-10,0] & U[0,50] & U[0,50] & 1   & 1 \\
    CM2 & U[-10,0] & U[0,100] & U[0,50] & 1   & 1 \\
    CM3 & U[-10,0] & U[0,50] & U[0,100] & 1   & 1 \\
    CM4 & U[-10,0] & U[0,50] & U[0,50] & U[0,30] & U[0,30] \\
    \bottomrule
    \end{tabular*}%
  \label{t3}%
\end{table*}%
In this section, the results of the MOBA model that accounted for the costs are evaluated in comparison to those from the model by Tortorell in which an ROC convex hull (ROCCH) curve was constructed from the confidence scores and the rejection thresholds were determined based on the tangents of the ROCCH curve \cite{tortorella2004reducing,tortorella2000optimal,santos2005optimal}. In Tortorell's model, when the condition
\begin{equation}
\frac{CTN-CRN}{CFN-CRP} > \frac{CFP-CRN}{CTP-CRP}
\end{equation}
was not satisfied, the reject option could not be activated. That is to say, the traditional classifier without rejection could provide the minimal cost. In this experiment, the twin support vector machine (TWSVM) \cite{khemchandani2007twin} was used as the scoring classifier to generate confidence scores \cite{lin2017twin}. Four cost models \cite{lin2017twin} shown in Table~\ref{t3} were used, where U[a,b] denotes a uniform distribution over the interval [a,b]. Note that while CRP and CRN had equal values in CM4 in~\cite{lin2017twin}, in this experiment, CRP and CRN had different values in CM4 as class-dependent reject costs were considered. 
 
A Wilcoxon rank sum test \cite{lin2017twin,tortorella2004reducing} was performed to compare the two cost-related abstaining models. The details of the Wilcoxon rank sum test can be found in~\cite{lin2017twin}. In this test, 1000 cost matrices (CTP/N, CFP/N, and CRP/N) were generated for each cost model in Table~\ref{t3}. Then, for each cost matrix, the expected cost was computed as per Eq.~(\ref{eq3}). Finally, the numbers of cases where the cost of the MOBA model was lower, higher, or identical compared to the cost of Tortorell's model were counted. There were two scenarios that resulted in identical costs: 1) the costs of the compared methods were equal; or 2) for a certain cost matrix, the reject option in Tortorell's model was not activated, in which situation, no MOBA model was constructed. The hyperparameters in the MOBA model were set as follows. First, Tortorell's model was enforced and the reject rates with respect to the two classes were obtained. Then, the values corresponding to the reject rates were assigned to $p_{max}$ and $n_{max}$. The two reject rates obtained via Tortorell's model can be regarded as good candidates to avoid blindly setting the hyperparameters in the MOBA model. 
% table 4
\begin{table*}[!htb]
  \centering
  \caption{Results of the Wilcoxon rank sum test for comparing MOBA and Tortorell's model}
    \begin{tabular*}{\textwidth}{l@{\extracolsep{\fill}}cccclcccc}
    \toprule
        & CM1 & CM2 & CM3 & CM4 &     & \multicolumn{1}{c}{CM1} & \multicolumn{1}{c}{CM2} & \multicolumn{1}{c}{CM3} & \multicolumn{1}{c}{CM4} \\
    \midrule
    \multicolumn{1}{l}{ionosphere} & \textbf{744} & \textbf{749} & \textbf{781} & 471 & \multicolumn{1}{l}{haberman} & \multicolumn{1}{c}{\textbf{466}} & \multicolumn{1}{c}{393} & \multicolumn{1}{c}{\textbf{523}} & \multicolumn{1}{c}{\textbf{484}} \\
        & 158 & 203 & 182 & 56  &     & \multicolumn{1}{c}{436} & \multicolumn{1}{c}{\textbf{559}} & \multicolumn{1}{c}{440} & \multicolumn{1}{c}{43} \\
        & 98  & 48  & 37  & \textbf{473} &     & \multicolumn{1}{c}{98} & \multicolumn{1}{c}{48} & \multicolumn{1}{c}{37} & \multicolumn{1}{c}{473} \\
    \multicolumn{1}{l}{pima} & \textbf{459} & \textbf{597} & \textbf{524} & 412 & \multicolumn{1}{l}{cmc} & \multicolumn{1}{c}{93} & \multicolumn{1}{c}{45} & \multicolumn{1}{c}{135} & \multicolumn{1}{c}{338} \\
        & 443 & 350 & 439 & 115 &     & \multicolumn{1}{c}{65} & \multicolumn{1}{c}{23} & \multicolumn{1}{c}{67} & \multicolumn{1}{c}{57} \\
        & 98  & 53  & 37  & \textbf{473} &     & \multicolumn{1}{c}{\textbf{842}} & \multicolumn{1}{c}{\textbf{932}} & \multicolumn{1}{c}{\textbf{798}} & \multicolumn{1}{c}{\textbf{605}} \\
    \multicolumn{1}{l}{credit-g} & \textbf{489} & \textbf{516} & 480 & 444 & \multicolumn{1}{l}{transfusion} & \multicolumn{1}{c}{\textbf{621}} & \multicolumn{1}{c}{\textbf{619}} & \multicolumn{1}{c}{\textbf{529}} & \multicolumn{1}{c}{\textbf{484}} \\
        & 413 & 436 & \textbf{483} & 83  &     & \multicolumn{1}{c}{281} & \multicolumn{1}{c}{333} & \multicolumn{1}{c}{434} & \multicolumn{1}{c}{43} \\
        & 98  & 48  & 37  & \textbf{473} &     & \multicolumn{1}{c}{98} & \multicolumn{1}{c}{48} & \multicolumn{1}{c}{37} & \multicolumn{1}{c}{473} \\
    \multicolumn{1}{l}{ecoli3} & \textbf{738} & \textbf{531} & \textbf{522} & 438 & \multicolumn{1}{l}{Australian} & \multicolumn{1}{c}{\textbf{659}} & \multicolumn{1}{c}{\textbf{711}} & \multicolumn{1}{c}{\textbf{664}} & \multicolumn{1}{c}{\textbf{473}} \\
        & 164 & 416 & 441 & 89  &     & \multicolumn{1}{c}{243} & \multicolumn{1}{c}{241} & \multicolumn{1}{c}{299} & \multicolumn{1}{c}{54} \\
        & 98  & 53  & 37  & \textbf{473} &     & \multicolumn{1}{c}{98} & \multicolumn{1}{c}{48} & \multicolumn{1}{c}{37} & \multicolumn{1}{c}{\textbf{473}} \\
    \multicolumn{1}{l}{hepatitis} & \textbf{510} & \textbf{520} & \textbf{514} & 435 &     &     &     &     &  \\
        & 392 & 432 & 449 & 92  &     &     &     &     &  \\
        & 98  & 48  & 37  & \textbf{473} &     &     &     &     &  \\
    \bottomrule
    \end{tabular*}%
  \label{t4}%
\end{table*}%

The results of the Wilcoxon rank sum test are shown in Table~\ref{t4}. Note that in each scenario (each dataset in each cost model), the three figures from top to bottom represent the numbers of lower, higher, and identical costs in the MOBA model compared to the costs obtained via Tortorell's model, respectively. In the table, it can be seen that there were a greater number of lower-cost cases for the MOBA model in almost all of the scenarios in the CM1, CM2, and CM3 cost models, while for most scenarios in the CM4 cost model, the identical-cost case was the most frequent. This is because the number of inactivated reject options increased. In the remaining two cases (i.e., in the lower- and higher-cost cases), the MOBA model provided a lower cost than Tortorell's model in the vast majority of cases. In addition, for a certain cost matrix, the reject option may not be activated in Tortorell's model, whereas the rejection could still be enforced in the MOBA model. This is an advantage of the MOBA model over Tortorell's model; that is, the MOBA model is not subject to a certain cost matrix.

\subsection{Comparison of the results when the costs are not considered}
\label{sec4.2}
\begin{figure*}[!htb]
\centering
\subfigure[ACC-Rej]{
\includegraphics[width=0.3\textwidth]{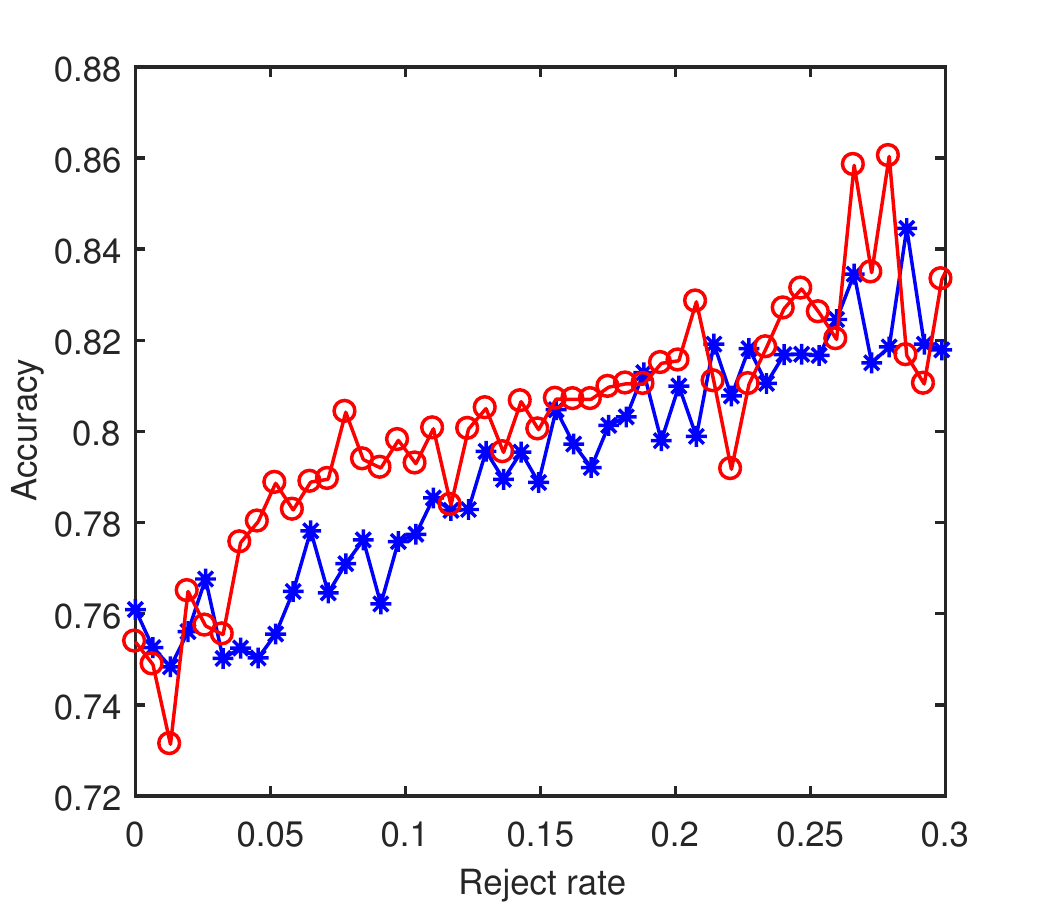}
}
\subfigure[AUC-Rej]{
\includegraphics[width=0.3\textwidth]{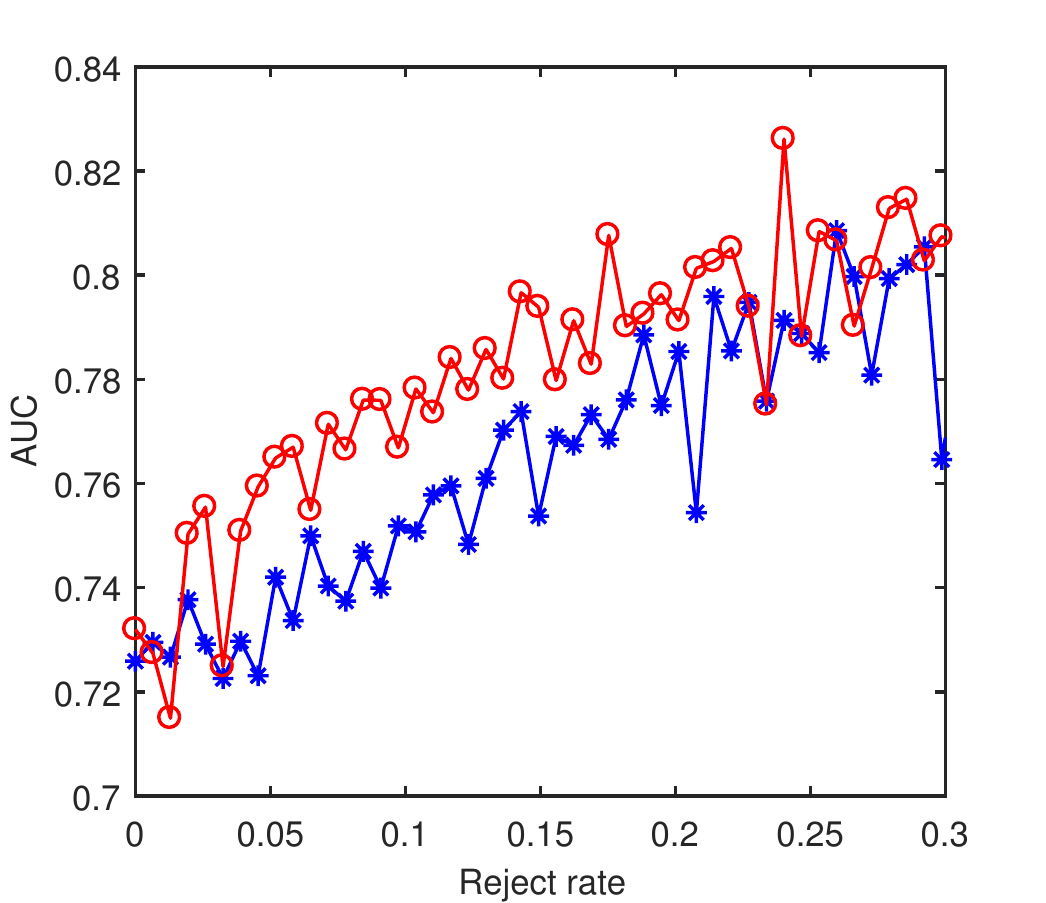}
}
\subfigure[G-Rej]{
\includegraphics[width=0.3\textwidth]{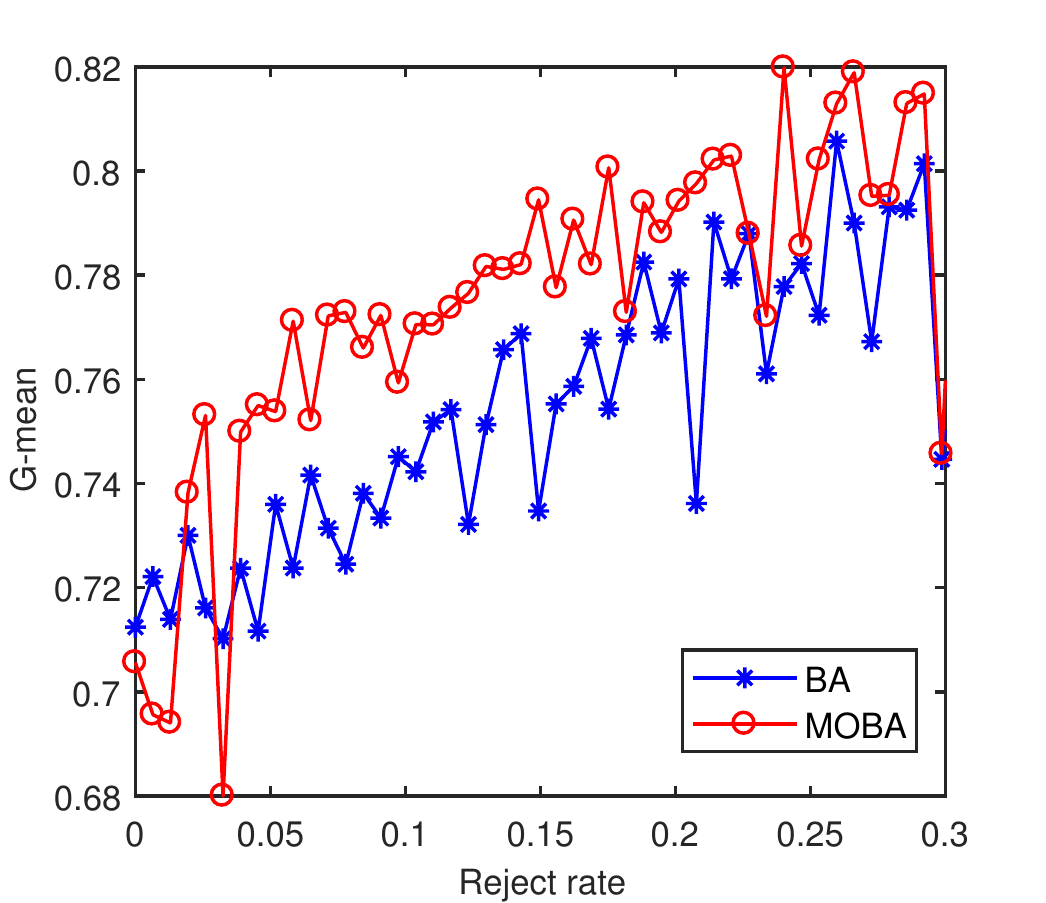}
}
\caption{Performance-rejection curves of BA and MOBA models using \textit{pima} dataset}\label{f3}
\end{figure*}

\begin{figure*}[!htb]
\centering
\subfigure[ACC-Rej]{
\includegraphics[width=0.3\textwidth]{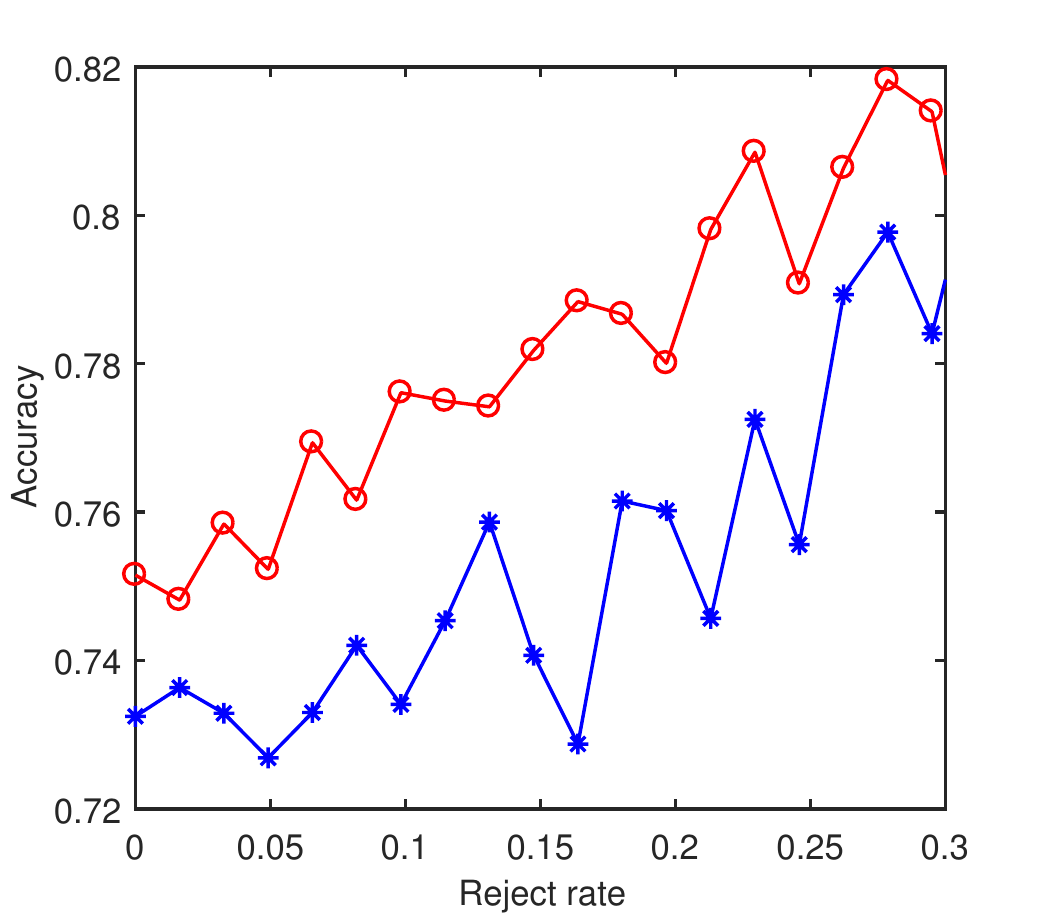}
}
\subfigure[AUC-Rej]{
\includegraphics[width=0.3\textwidth]{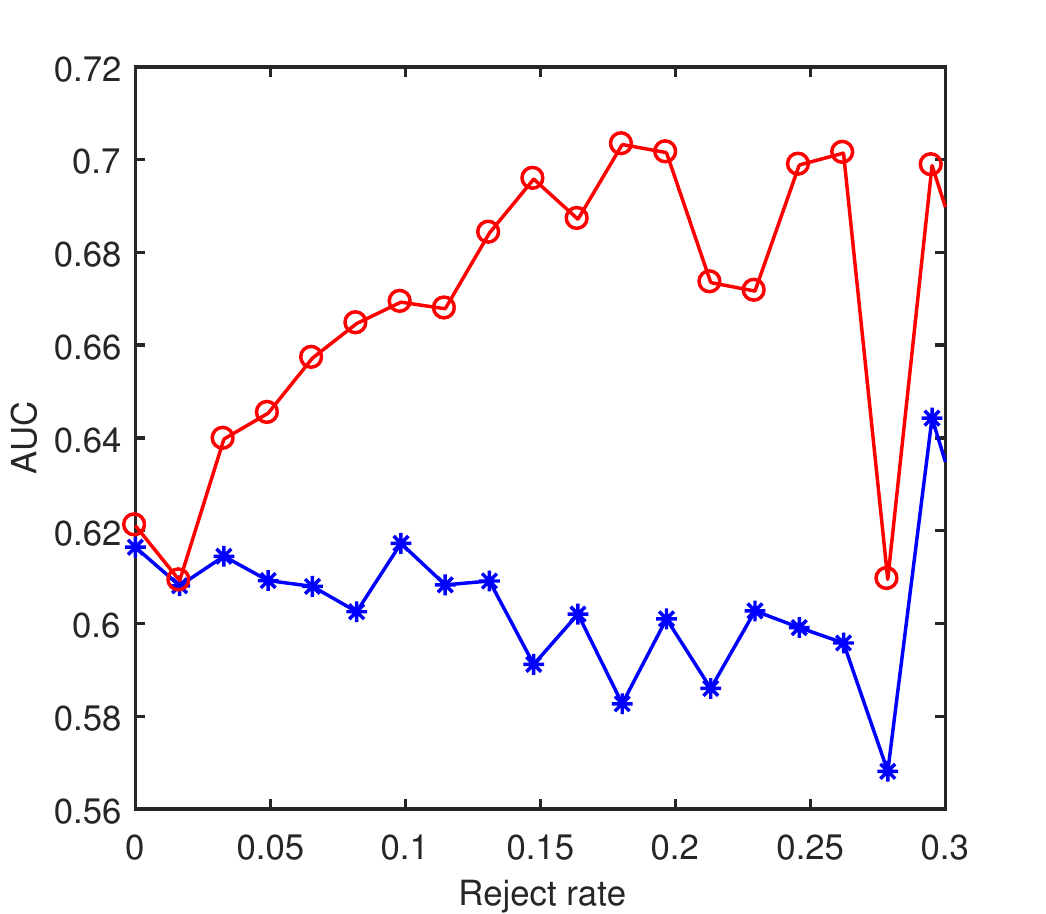}
}
\subfigure[G-Rej]{
\includegraphics[width=0.3\textwidth]{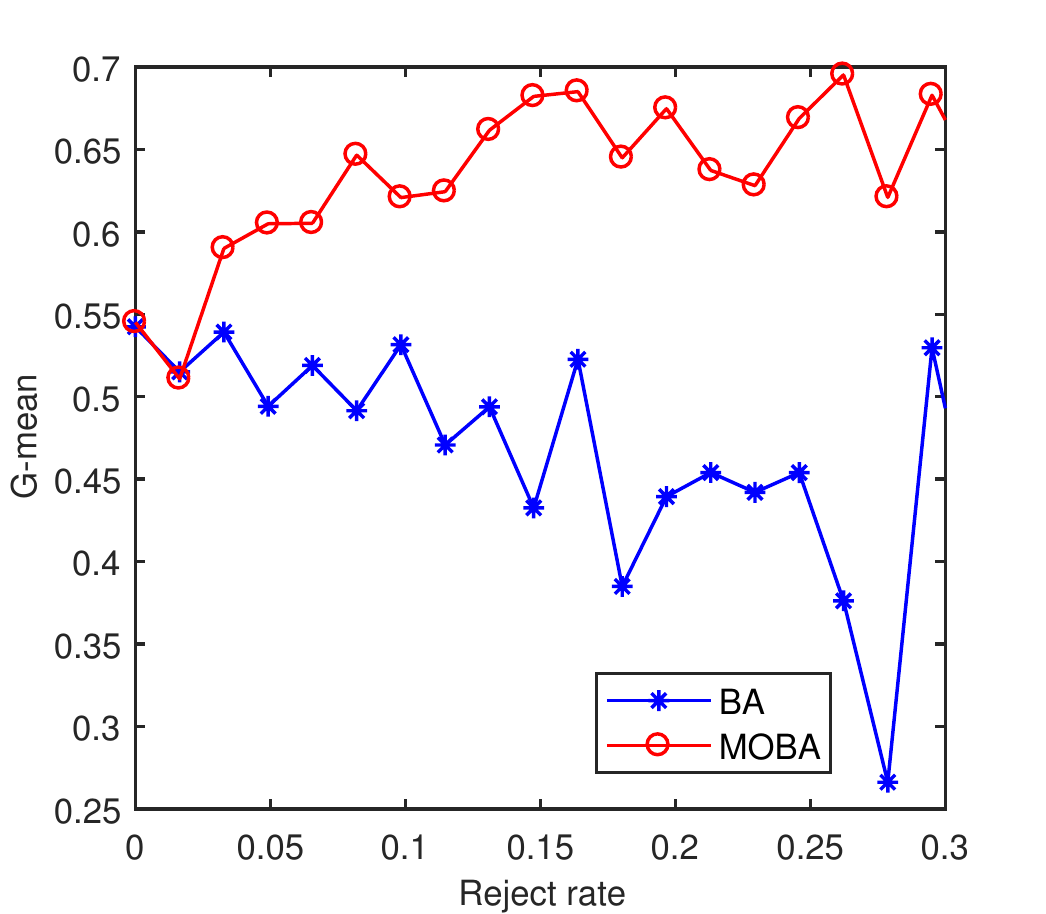}
}
\caption{Performance-rejection curves of BA and MOBA models using \textit{haberman} dataset}\label{f4}
\end{figure*}

\begin{figure*}[!htb]
\centering
\subfigure[ACC-Rej]{
\includegraphics[width=0.3\textwidth]{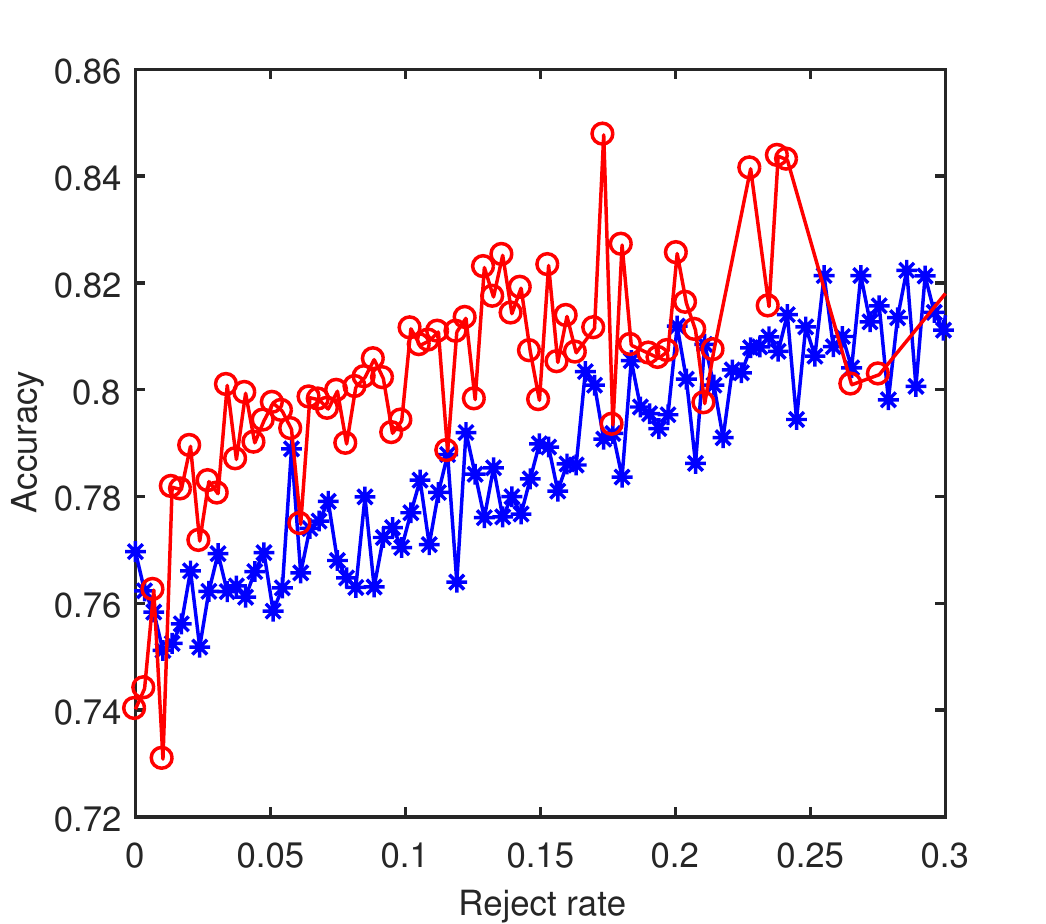}
}
\subfigure[AUC-Rej]{
\includegraphics[width=0.3\textwidth]{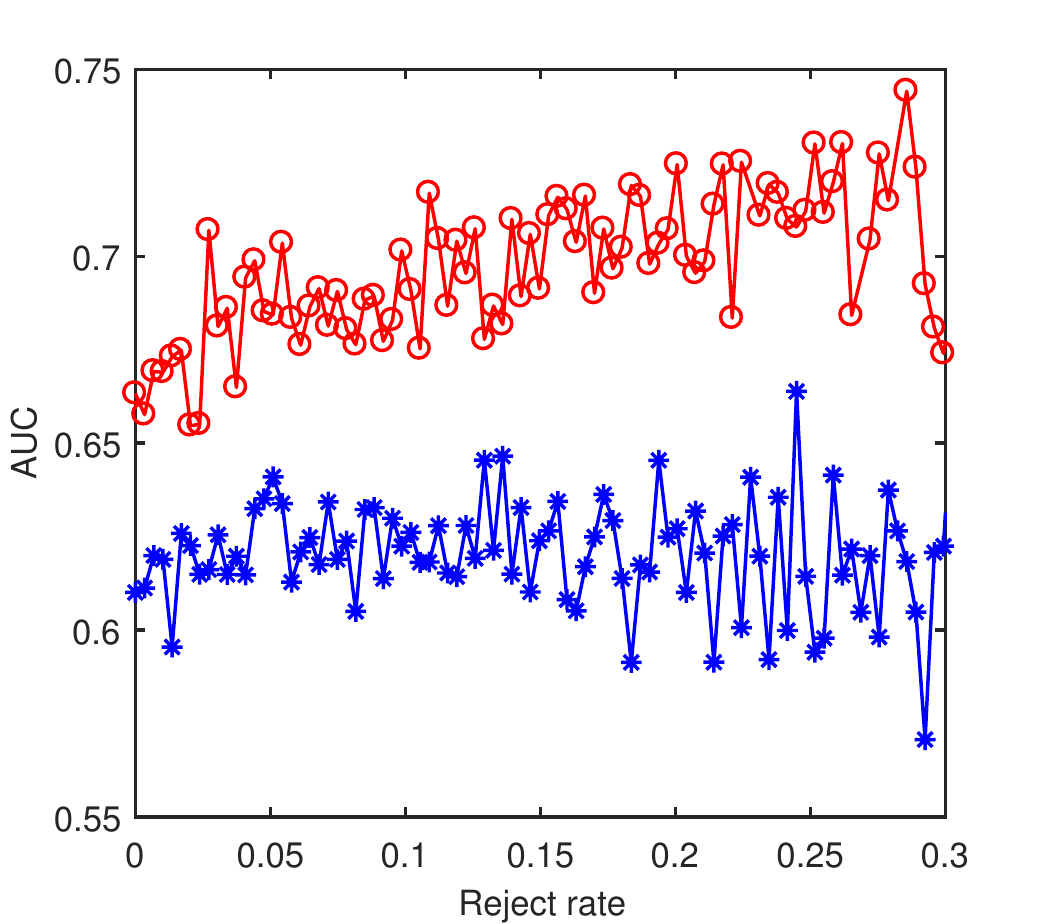}
}
\subfigure[G-Rej]{
\includegraphics[width=0.3\textwidth]{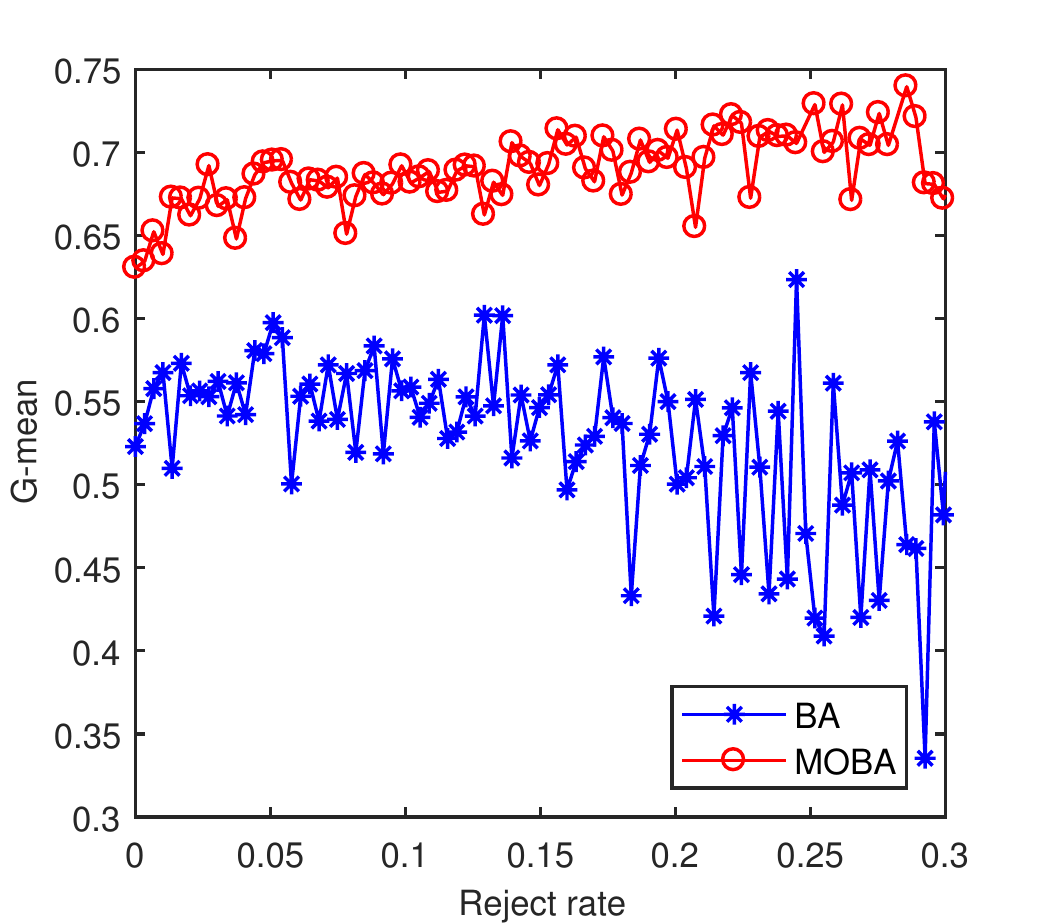}
}
\caption{Performance-rejection curves of BA and MOBA models using \textit{cmc} dataset}\label{f5}
\end{figure*}

\begin{figure*}[!htb]
\centering
\subfigure[ACC-Rej]{
\includegraphics[width=0.3\textwidth]{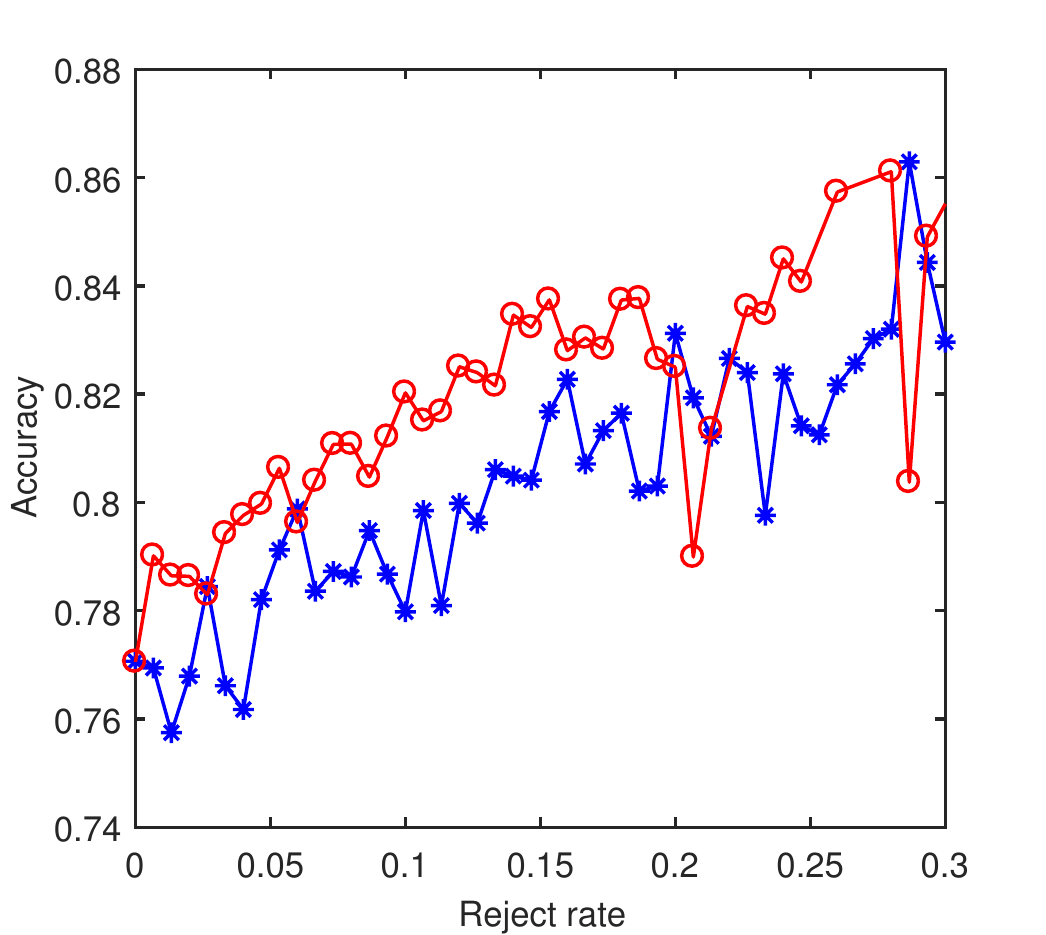}
}
\subfigure[AUC-Rej]{
\includegraphics[width=0.3\textwidth]{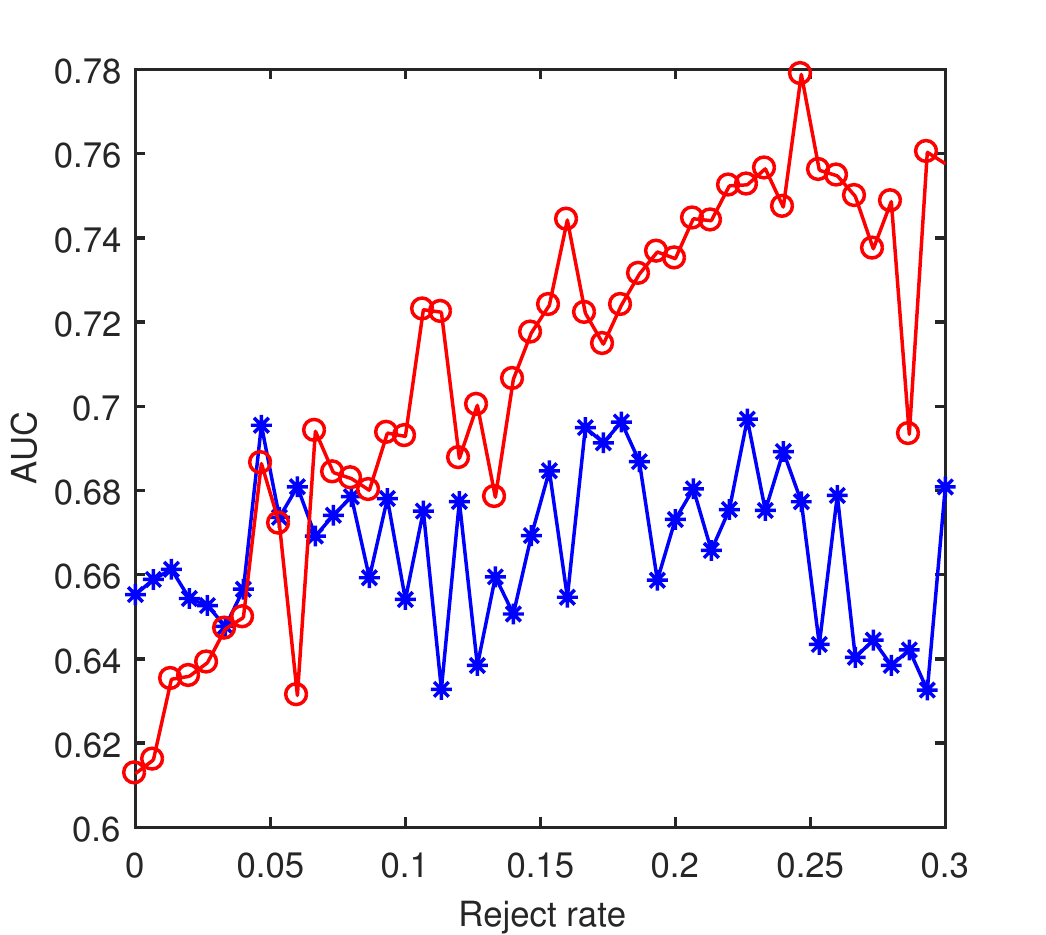}
}
\subfigure[G-Rej]{
\includegraphics[width=0.3\textwidth]{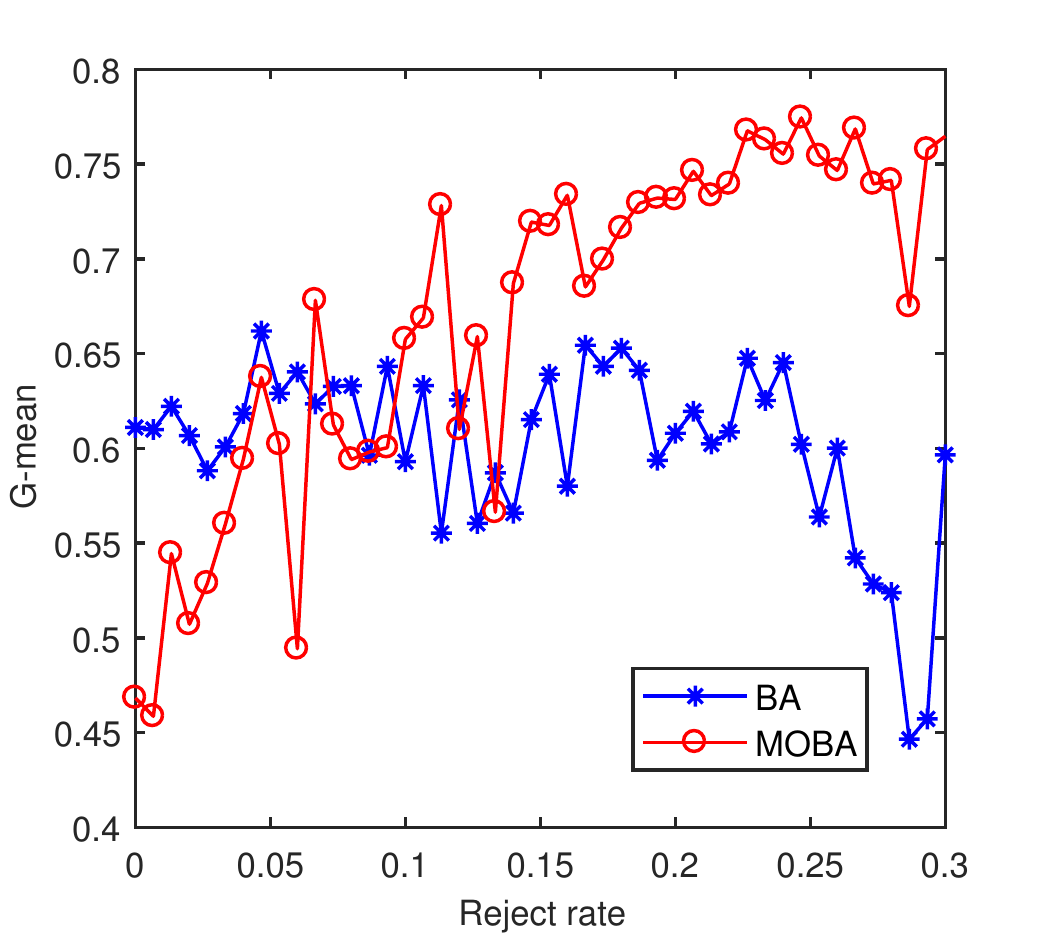}
}
\caption{The performance-rejection curves of BA and MOBA models using \textit{transfusion} dataset}\label{f6}
\end{figure*}

In this section, the results of the MOBA model that did not consider cost information are presented in comparison to those of the BA model \cite{pietraszek2007use}. In the BA model, the cost ratio between CFN to CFP was set to one, which means that the CFN and CFP were the same. For this reason, the values of $p_{max}$ and $n_{max}$ in the MOBA model were set to the same values. When $p_{max} = n_{max}$, the overall reject rate $k_{max}$ in (\ref{eq4}) was equal to the reject rate of each class ($p_{max}$ or $n_{max}$), which ensures the comparability of the BA and MOBA models. The abstention parameters were set from 0.01 to 0.3 in steps of 0.02 since larger rejection rates are usually of no significance in practical applications \cite{simeone2012design}. For models that do not consider the costs, the classification results are typically evaluated based on the corresponding performance-rejection curves. Here, the accuracy (ACC), AUC, and G-mean (G) were used as the evaluation metrics. Note that the G-mean is the geometric mean of the sensitivity and specificity. That is, ACC-Rej, AUC-Rej, and G-Rej curves were obtained using each dataset. In this paper, only the trade-off curves of datasets \textit{pima}, \textit{haberman}, \textit{cmc}, and \textit{transfusion} are shown (Figures~\ref{f3}-\ref{f6}) to discuss the performance of the two compared models. Similar trade-off curves were obtained for the other datasets.


Overall, the MOBA model provided a better trade-off between the performance (accuracy, AUC, and G-mean) and rejection than the BA model. In other words, for a fixed reject rate, the MOBA model provided higher values for the accuracy, AUC, and G-mean than those of the BA model. For the MOBA model, the values of the accuracy, AUC, and G-mean grew with increasing reject rates, while for the BA model, only the ACC-Rej curves exhibited an increasing trend and the AUC and G-mean values decreased in Figures~\ref{f4}-\ref{f6}. This is because the BA model minimized the error rate under an overall reject rate and did not consider the class-dependent performance. According to the definitions, the values of the AUC and G-mean were large only when both sensitivity and specificity were large. Thus, the MOBA model was shown to provide more balanced sensitivity and specificity values. The fluctuations observed in the trade-off curves may be attributed to the small sample size and randomness when constructing the models. 

\section{Conclusions} \label{sec5}
In this paper, the MOBA model was proposed for cost-sensitive problems where the error costs are usually unknown and asymmetric. The MOBA model avoids setting cost information by optimizing essential metrics. A significant advantage of the MOBA model is its robustness towards different conditions, such as known or unknown cost information, evolving costs over time, and using different metrics to evaluate performance. Also, the MOBA model can accommodate unbalanced classification problems due to its ability to optimize the respective performance of two classes under class-dependent abstention constraints \cite{he2008learning,xiao2017class}. Experimental results in this study have shown that the MOBA model performed better than previous models. When the costs were known, the MOBA model obtained a greater number of lower-cost cases in the Wilcoxon rank sum test than did Tortorell's model. When costs were not considered, the MOBA model achieved better trade-offs between performance (accuracy, AUC, and G-mean) and abstention than the BA model. 

Abstaining classification has wide applications in safety-critical fields, such as medical diagnosis, fault detection, credit assessment, and so on. Rejection means it is difficult to make a definite decision given the current knowledge. More knowledge should be provided to reduce misclassification costs. The proposed MOBA model will show a promising utility owing to its advantages and better performance. In the future, we intend to extend the MOBA model to multi-class problems and apply the MOBA model to specific applications.

\section*{Declarations of interest}
None.

\section*{References}
\bibliographystyle{model1-num-names}
\bibliography{sample}

\begin{thebibliography}{28}
\expandafter\ifx\csname natexlab\endcsname\relax\def\natexlab#1{#1}\fi
\providecommand{\url}[1]{\texttt{#1}}
\providecommand{\href}[2]{#2}
\providecommand{\path}[1]{#1}
\providecommand{\DOIprefix}{doi:}
\providecommand{\ArXivprefix}{arXiv:}
\providecommand{\URLprefix}{URL: }
\providecommand{\Pubmedprefix}{pmid:}
\providecommand{\doi}[1]{\href{http://dx.doi.org/#1}{\path{#1}}}
\providecommand{\Pubmed}[1]{\href{pmid:#1}{\path{#1}}}
\providecommand{\bibinfo}[2]{#2}
\ifx\xfnm\relax \def\xfnm[#1]{\unskip,\space#1}\fi
%Type = Article
\bibitem[{Chen et~al.(2018)Chen, Wang, Hao, and Zhao}]{chen2018evidential}
\bibinfo{author}{X.~Chen}, \bibinfo{author}{P.~Wang}, \bibinfo{author}{Y.~Hao},
  \bibinfo{author}{M.~Zhao},
\newblock \bibinfo{title}{Evidential {KNN}-based condition monitoring and early
  warning method with applications in power plant},
\newblock \bibinfo{journal}{Neurocomputing}  (\bibinfo{year}{2018}).
%Type = Article
\bibitem[{Kang et~al.(2017)Kang, Cho, Rhee, and Yu}]{kang2017reliable}
\bibinfo{author}{S.~Kang}, \bibinfo{author}{S.~Cho}, \bibinfo{author}{S.-j.
  Rhee}, \bibinfo{author}{K.-S. Yu},
\newblock \bibinfo{title}{Reliable prediction of anti-diabetic drug failure
  using a reject option},
\newblock \bibinfo{journal}{Pattern Analysis and Applications}
  \bibinfo{volume}{20} (\bibinfo{year}{2017}) \bibinfo{pages}{883--891}.
%Type = Article
\bibitem[{Lin et~al.(2018)Lin, Sun, Toh, Zhang, and Lin}]{lin2018biomedical}
\bibinfo{author}{D.~Lin}, \bibinfo{author}{L.~Sun}, \bibinfo{author}{K.-A.
  Toh}, \bibinfo{author}{J.~B. Zhang}, \bibinfo{author}{Z.~Lin},
\newblock \bibinfo{title}{Biomedical image classification based on a cascade of
  an svm with a reject option and subspace analysis},
\newblock \bibinfo{journal}{Computers in biology and medicine}
  \bibinfo{volume}{96} (\bibinfo{year}{2018}) \bibinfo{pages}{128--140}.
%Type = Article
\bibitem[{Wang et~al.(2017)Wang, Wang, He, Gu, and Yan}]{wang2017fault}
\bibinfo{author}{Z.~Wang}, \bibinfo{author}{Z.~Wang}, \bibinfo{author}{S.~He},
  \bibinfo{author}{X.~Gu}, \bibinfo{author}{Z.~F. Yan},
\newblock \bibinfo{title}{Fault detection and diagnosis of chillers using
  bayesian network merged distance rejection and multi-source non-sensor
  information},
\newblock \bibinfo{journal}{Applied energy} \bibinfo{volume}{188}
  (\bibinfo{year}{2017}) \bibinfo{pages}{200--214}.
%Type = Article
\bibitem[{Chow(1970)}]{chow1970optimum}
\bibinfo{author}{C.~Chow},
\newblock \bibinfo{title}{On optimum recognition error and reject tradeoff},
\newblock \bibinfo{journal}{IEEE Transactions on information theory}
  \bibinfo{volume}{16} (\bibinfo{year}{1970}) \bibinfo{pages}{41--46}.
%Type = Article
\bibitem[{Chow(1957)}]{chow1957optimum}
\bibinfo{author}{C.-K. Chow},
\newblock \bibinfo{title}{An optimum character recognition system using
  decision functions},
\newblock \bibinfo{journal}{IRE Transactions on Electronic Computers}
  (\bibinfo{year}{1957}) \bibinfo{pages}{247--254}.
%Type = Article
\bibitem[{Tortorella(2004)}]{tortorella2004reducing}
\bibinfo{author}{Tortorella},
\newblock \bibinfo{title}{Reducing the classification cost of support vector
  classifiers through an {ROC}-based reject rule},
\newblock \bibinfo{journal}{Pattern Analysis and Applications}
  \bibinfo{volume}{7} (\bibinfo{year}{2004}) \bibinfo{pages}{128--143}.
%Type = Inproceedings
\bibitem[{Tortorella(2000)}]{tortorella2000optimal}
\bibinfo{author}{F.~Tortorella},
\newblock \bibinfo{title}{An optimal reject rule for binary classifiers},
\newblock in: \bibinfo{booktitle}{Joint IAPR International Workshops on
  Statistical Techniques in Pattern Recognition (SPR) and Structural and
  Syntactic Pattern Recognition (SSPR)}, \bibinfo{organization}{Springer},
  \bibinfo{year}{2000}, pp. \bibinfo{pages}{611--620}.
%Type = Article
\bibitem[{Pietraszek(2007)}]{pietraszek2007use}
\bibinfo{author}{T.~Pietraszek},
\newblock \bibinfo{title}{On the use of {ROC} analysis for the optimization of
  abstaining classifiers},
\newblock \bibinfo{journal}{Machine Learning} \bibinfo{volume}{68}
  (\bibinfo{year}{2007}) \bibinfo{pages}{137--169}.
%Type = Article
\bibitem[{Vanderlooy et~al.(2009)Vanderlooy, Sprinkhuizen-Kuyper, Smirnov, and
  van~den Herik}]{vanderlooy2009roc}
\bibinfo{author}{S.~Vanderlooy}, \bibinfo{author}{I.~G. Sprinkhuizen-Kuyper},
  \bibinfo{author}{E.~N. Smirnov}, \bibinfo{author}{H.~J. van~den Herik},
\newblock \bibinfo{title}{The {ROC} isometrics approach to construct reliable
  classifiers},
\newblock \bibinfo{journal}{Intelligent Data Analysis} \bibinfo{volume}{13}
  (\bibinfo{year}{2009}) \bibinfo{pages}{3--37}.
%Type = Article
\bibitem[{Deb et~al.(2002)Deb, Pratap, Agarwal, and Meyarivan}]{deb2002fast}
\bibinfo{author}{K.~Deb}, \bibinfo{author}{A.~Pratap},
  \bibinfo{author}{S.~Agarwal}, \bibinfo{author}{T.~Meyarivan},
\newblock \bibinfo{title}{A fast and elitist multiobjective genetic algorithm:
  {NSGA-II}},
\newblock \bibinfo{journal}{IEEE transactions on evolutionary computation}
  \bibinfo{volume}{6} (\bibinfo{year}{2002}) \bibinfo{pages}{182--197}.
%Type = Article
\bibitem[{Zitzler et~al.(2000)Zitzler, Deb, and Thiele}]{zitzler2000comparison}
\bibinfo{author}{E.~Zitzler}, \bibinfo{author}{K.~Deb},
  \bibinfo{author}{L.~Thiele},
\newblock \bibinfo{title}{Comparison of multiobjective evolutionary algorithms:
  Empirical results},
\newblock \bibinfo{journal}{Evolutionary computation} \bibinfo{volume}{8}
  (\bibinfo{year}{2000}) \bibinfo{pages}{173--195}.
%Type = Article
\bibitem[{Zitzler et~al.(2001)Zitzler, Laumanns, and Thiele}]{zitzler2001spea2}
\bibinfo{author}{E.~Zitzler}, \bibinfo{author}{M.~Laumanns},
  \bibinfo{author}{L.~Thiele},
\newblock \bibinfo{title}{{SPEA}2: Improving the strength pareto evolutionary
  algorithm},
\newblock \bibinfo{journal}{TIK-report} \bibinfo{volume}{103}
  (\bibinfo{year}{2001}).
%Type = Inproceedings
\bibitem[{Corne et~al.(2001)Corne, Jerram, Knowles, and Oates}]{corne2001pesa}
\bibinfo{author}{D.~W. Corne}, \bibinfo{author}{N.~R. Jerram},
  \bibinfo{author}{J.~D. Knowles}, \bibinfo{author}{M.~J. Oates},
\newblock \bibinfo{title}{{PESA-II}: Region-based selection in evolutionary
  multiobjective optimization},
\newblock in: \bibinfo{booktitle}{Proceedings of the 3rd Annual Conference on
  Genetic and Evolutionary Computation}, \bibinfo{organization}{Morgan Kaufmann
  Publishers Inc.}, \bibinfo{year}{2001}, pp. \bibinfo{pages}{283--290}.
%Type = Article
\bibitem[{Srinivas and Deb(1994)}]{srinivas1994muiltiobjective}
\bibinfo{author}{N.~Srinivas}, \bibinfo{author}{K.~Deb},
\newblock \bibinfo{title}{Muiltiobjective optimization using nondominated
  sorting in genetic algorithms},
\newblock \bibinfo{journal}{Evolutionary computation} \bibinfo{volume}{2}
  (\bibinfo{year}{1994}) \bibinfo{pages}{221--248}.
%Type = Article
\bibitem[{Fawcett(2006)}]{fawcett2006introduction}
\bibinfo{author}{T.~Fawcett},
\newblock \bibinfo{title}{An introduction to roc analysis},
\newblock \bibinfo{journal}{Pattern recognition letters} \bibinfo{volume}{27}
  (\bibinfo{year}{2006}) \bibinfo{pages}{861--874}.
%Type = Article
\bibitem[{Agrawal et~al.(1995)Agrawal, Deb, and Agrawal}]{agrawal1995simulated}
\bibinfo{author}{R.~B. Agrawal}, \bibinfo{author}{K.~Deb},
  \bibinfo{author}{R.~Agrawal},
\newblock \bibinfo{title}{Simulated binary crossover for continuous search
  space},
\newblock \bibinfo{journal}{Complex systems} \bibinfo{volume}{9}
  (\bibinfo{year}{1995}) \bibinfo{pages}{115--148}.
%Type = Inproceedings
\bibitem[{Kakde(2004)}]{kakde2004survey}
\bibinfo{author}{M.~R.~O. Kakde},
\newblock \bibinfo{title}{Survey on multiobjective evolutionary and real coded
  genetic algorithms},
\newblock in: \bibinfo{booktitle}{Proceedings of the 8th Asia Pacific symposium
  on intelligent and evolutionary systems}, \bibinfo{organization}{Citeseer},
  \bibinfo{year}{2004}, pp. \bibinfo{pages}{150--161}.
%Type = Article
\bibitem[{Beyer and Deb(2001)}]{beyer2001self}
\bibinfo{author}{H.-G. Beyer}, \bibinfo{author}{K.~Deb},
\newblock \bibinfo{title}{On self-adaptive features in real-parameter
  evolutionary algorithms},
\newblock \bibinfo{journal}{IEEE Transactions on evolutionary computation}
  \bibinfo{volume}{5} (\bibinfo{year}{2001}) \bibinfo{pages}{250--270}.
%Type = Article
\bibitem[{L{\'o}pez et~al.(2013)L{\'o}pez, Fern{\'a}ndez, Garc{\'\i}a, Palade,
  and Herrera}]{lopez2013insight}
\bibinfo{author}{V.~L{\'o}pez}, \bibinfo{author}{A.~Fern{\'a}ndez},
  \bibinfo{author}{S.~Garc{\'\i}a}, \bibinfo{author}{V.~Palade},
  \bibinfo{author}{F.~Herrera},
\newblock \bibinfo{title}{An insight into classification with imbalanced data:
  Empirical results and current trends on using data intrinsic
  characteristics},
\newblock \bibinfo{journal}{Information Sciences} \bibinfo{volume}{250}
  (\bibinfo{year}{2013}) \bibinfo{pages}{113--141}.
%Type = Article
\bibitem[{Pietraszek(2007)}]{pietraszek2007classification}
\bibinfo{author}{T.~Pietraszek},
\newblock \bibinfo{title}{Classification of intrusion detection alerts using
  abstaining classifiers},
\newblock \bibinfo{journal}{Intelligent Data Analysis} \bibinfo{volume}{11}
  (\bibinfo{year}{2007}) \bibinfo{pages}{293--316}.
%Type = Article
\bibitem[{Alcal{\'a}-Fdez et~al.(2011)Alcal{\'a}-Fdez, Fern{\'a}ndez, Luengo,
  Derrac, Garc{\'\i}a, S{\'a}nchez, and Herrera}]{alcala2011keel}
\bibinfo{author}{J.~Alcal{\'a}-Fdez}, \bibinfo{author}{A.~Fern{\'a}ndez},
  \bibinfo{author}{J.~Luengo}, \bibinfo{author}{J.~Derrac},
  \bibinfo{author}{S.~Garc{\'\i}a}, \bibinfo{author}{L.~S{\'a}nchez},
  \bibinfo{author}{F.~Herrera},
\newblock \bibinfo{title}{Keel data-mining software tool: data set repository,
  integration of algorithms and experimental analysis framework.},
\newblock \bibinfo{journal}{Journal of Multiple-Valued Logic \& Soft Computing}
  \bibinfo{volume}{17} (\bibinfo{year}{2011}).
%Type = Article
\bibitem[{Santos-Pereira and Pires(2005)}]{santos2005optimal}
\bibinfo{author}{C.~M. Santos-Pereira}, \bibinfo{author}{A.~M. Pires},
\newblock \bibinfo{title}{On optimal reject rules and roc curves},
\newblock \bibinfo{journal}{Pattern recognition letters} \bibinfo{volume}{26}
  (\bibinfo{year}{2005}) \bibinfo{pages}{943--952}.
%Type = Article
\bibitem[{Khemchandani et~al.(2007)Khemchandani, Chandra
  et~al.}]{khemchandani2007twin}
\bibinfo{author}{R.~Khemchandani}, \bibinfo{author}{S.~Chandra}, et~al.,
\newblock \bibinfo{title}{Twin support vector machines for pattern
  classification},
\newblock \bibinfo{journal}{IEEE Transactions on pattern analysis and machine
  intelligence} \bibinfo{volume}{29} (\bibinfo{year}{2007})
  \bibinfo{pages}{905--910}.
%Type = Article
\bibitem[{Lin et~al.(2017)Lin, Sun, Toh, Zhang, and Lin}]{lin2017twin}
\bibinfo{author}{D.~Lin}, \bibinfo{author}{L.~Sun}, \bibinfo{author}{K.-A.
  Toh}, \bibinfo{author}{J.~B. Zhang}, \bibinfo{author}{Z.~Lin},
\newblock \bibinfo{title}{Twin {SVM} with a reject option through roc curve},
\newblock \bibinfo{journal}{Journal of the Franklin Institute}
  (\bibinfo{year}{2017}).
%Type = Article
\bibitem[{Simeone et~al.(2012)Simeone, Marrocco, and
  Tortorella}]{simeone2012design}
\bibinfo{author}{P.~Simeone}, \bibinfo{author}{C.~Marrocco},
  \bibinfo{author}{F.~Tortorella},
\newblock \bibinfo{title}{Design of reject rules for ecoc classification
  systems},
\newblock \bibinfo{journal}{Pattern Recognition} \bibinfo{volume}{45}
  (\bibinfo{year}{2012}) \bibinfo{pages}{863--875}.
%Type = Article
\bibitem[{He and Garcia(2008)}]{he2008learning}
\bibinfo{author}{H.~He}, \bibinfo{author}{E.~A. Garcia},
\newblock \bibinfo{title}{Learning from imbalanced data},
\newblock \bibinfo{journal}{IEEE Transactions on Knowledge \& Data Engineering}
   (\bibinfo{year}{2008}) \bibinfo{pages}{1263--1284}.
%Type = Article
\bibitem[{Xiao et~al.(2017)Xiao, Zhang, Li, Zhang, and Yang}]{xiao2017class}
\bibinfo{author}{W.~Xiao}, \bibinfo{author}{J.~Zhang}, \bibinfo{author}{Y.~Li},
  \bibinfo{author}{S.~Zhang}, \bibinfo{author}{W.~Yang},
\newblock \bibinfo{title}{Class-specific cost regulation extreme learning
  machine for imbalanced classification},
\newblock \bibinfo{journal}{Neurocomputing} \bibinfo{volume}{261}
  (\bibinfo{year}{2017}) \bibinfo{pages}{70--82}.

\end{thebibliography}

\end{document}